\pdfoutput=1
\documentclass[11pt]{article}
\usepackage[colorlinks,bookmarksopen,bookmarksnumbered,citecolor=green, linkcolor=red, urlcolor=green]{hyperref}
\usepackage{xcolor}

\usepackage{tcolorbox}
\usepackage{amsmath}
\usepackage{amsfonts}
\usepackage{algorithm}
\usepackage{algpseudocode}
\usepackage{tcolorbox}

\usepackage[preprint]{acl}

\usepackage{times}
\usepackage{latexsym}
\usepackage{multirow}
\tcbuselibrary{skins, breakable, theorems}
\usepackage[T1]{fontenc}
\usepackage[utf8]{inputenc}

\usepackage{microtype}

\usepackage{inconsolata}

\usepackage{graphicx}

\title{LLM-empowered Dynamic Prompt Routing for Vision-Language Models Tuning under Long-Tailed Distributions}

\author{
Yongju Jia\textsuperscript{1} \quad
Jiarui Ma\textsuperscript{1} \quad
Xiangxian Li\textsuperscript{1}\thanks{Corresponding author.} \quad
Baiqiao Zhang\textsuperscript{1,2} \quad
Xianhui Cao\textsuperscript{3} \AND
Juan Liu\textsuperscript{1,4} \quad
Yulong Bian\textsuperscript{1,4} \\
\textsuperscript{1}Shandong University, Weihai, China \\
\textsuperscript{2}The Hong Kong University of Science and Technology, Hong Kong, China \\
\textsuperscript{3}AiLF Instruments, Weihai, China \\
\textsuperscript{4}Shandong Key Laboratory of Intelligent Electronic Packaging Testing and Application, Weihai, China \\
\texttt{jyjia@mail.sdu.edu.cn},  \texttt{jrma@mail.sdu.edu.cn}, \texttt{xiangxianli@sdu.edu.cn}, \\
\texttt{zzzliujuan@sdu.edu.cn}, \texttt{bianyulong@sdu.edu.cn}, \\
\texttt{baiqiao.zhang@connect.ust.hk}, \texttt{hans@ailf.com.cn}
}
\begin{document}
\maketitle

\begin{abstract}
Pre-trained vision-language models (VLMs), such as CLIP, have demonstrated impressive capability in visual tasks, but their fine-tuning often suffers from bias in class-imbalanced scene. Recent works have introduced large language models (LLMs) to enhance VLM fine-tuning with supplementing semantic information. However, they often overlook inherent class imbalance in VLMs' pre-training, which may lead to bias accumulation in downstream tasks. To address this problem, this paper proposes a Multi-dimensional Dynamic Prompt Routing (MDPR) framework. MDPR constructs a comprehensive knowledge base for classes, spanning five visual-semantic dimensions. During fine-tuning, the dynamic routing mechanism aligns global visual classes, retrieves optimal prompts, and balances fine-grained semantics, yielding stable predictions through logits fusion. Extensive experiments on long-tailed benchmarks, including CIFAR-LT, ImageNet-LT, and Places-LT, demonstrate that MDPR achieves comparable results with current SOTA methods. Ablation studies further confirm the effectiveness of our semantic library for tail classes, and show that our dynamic routing incurs minimal computational overhead, making MDPR a flexible and efficient enhancement for VLM fine-tuning under data imbalance.
\end{abstract}

\section{Introduction}
Pretrained Vision-Language Models (VLMs), such as CLIP \cite{radford2021learning}, have demonstrated remarkable capabilities in visual tasks by leveraging cross-modal knowledge with tuning techniques \cite{khattak2023maple,zhou2022conditional}. However, the fine-tuning of VLM under imbalanced downstream data exhibits significant bias \cite{wang2024exploring}, i.e., models favor many-sampled class optimization while under-performing on few-sampled classes, as shown in Figure 1(b) and (c). Lately, Large language Models (LLMs) are introduced to enhance VLM tuning, which faces two fundamental questions: (1) What semantic information from LLMs is effective in alleviating distributional bias? (2) How to leverage augmented information during the fine-tuning process?

\begin{figure}[t]
  \includegraphics[width=\columnwidth]{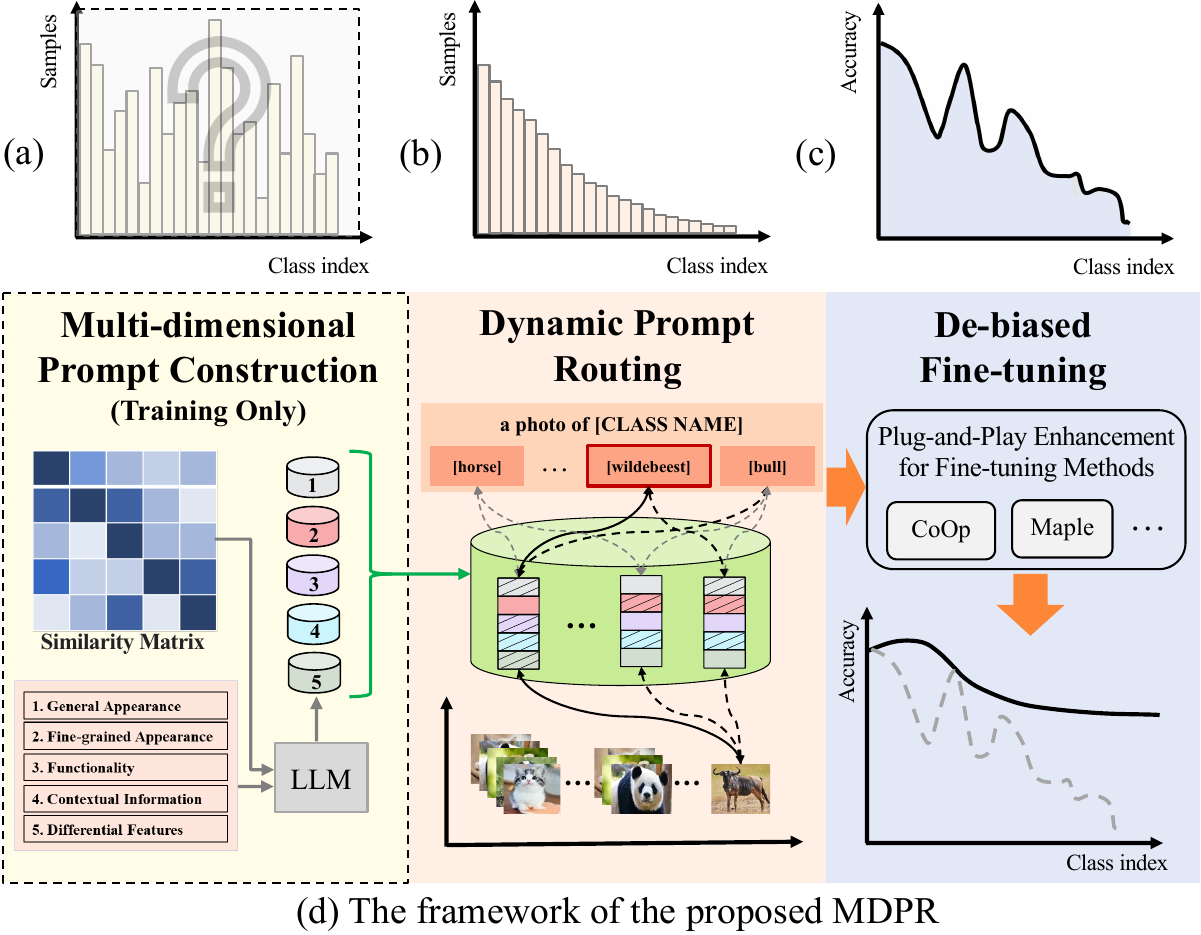}
  \caption{Illustration of how MDPR alleviate the bias. To address the (a) unknown imbalance in the pre-training of VLMs and (b) the long-tailed distribution in downstream data, which jointly lead the (c) accuracy bias in fine-tuning, (d) MDPR constructs comprehensive knowledge using offline LLM generation, and designs a dynamic prompt routing mechanism to enhancing fine-tuning methods with de-biasing the predictions.}
  \label{fig:motivation}
\end{figure}

% These studies about integrating LLMs into visual learning pipelines from several perspectives, majorly including class-level semantic enrichment, sample synthesis, and concept expansion. LLaMP \cite{zheng2024large} generates discriminative prompts for each class, LTGC \cite{zhao2024ltgc} guides image synthesis to balance class distributions, and PerVL \cite{cohen2022my} injects novel concepts through personalized descriptions. While these approaches demonstrate the potential of LLMs in enhancing visual representation learning, they still face notable limitations. Specifically, current methods often enhance downstream semantics without explicitly addressing the inherent bias embedded in pretrained VLMs \cite{zhu2023generalized,chen2025rethinking}, as illustrated in Figure 1(a), risking cumulative bias during fine-tuning. Moreover, many rely on static prompts, prior injection, or external generative models, lacking the adaptability to downstream task dynamics. 

To address VLMs' performance bottlenecks in data-scarce scenarios, prior works leverage LLMs for class-level semantic enhancement, sample synthesis, and open-world concept expansion. For semantic enhancement, LLMs generate discriminative prompts to improve inter-class separability \cite{zheng2024large}. For sample synthesis, LTGC \cite{zhao2024ltgc} guides diffusion models to synthesize tail-class samples. For concept expansion, PerVL \cite{cohen2022my} and Custom Diffusion \cite{kumari2023multi} enable open-set generalization via text descriptions. However, these methods often overlook intrinsic VLMs' biases, leading to cumulative bias during fine-tuning, and rely on static prompts or costly generative models, limiting adaptability. This necessitates an efficient framework for dynamic LLM-VLM interaction in fine-tuning.

To enhance the effectiveness of LLM knowledge in fine-tuning VLMs under imbalanced distributions, we propose Multi-dimensional Dynamic Prompt Routing (MDPR), as illustrated in Figure 1(d). Specifically, to address the implicit imbalance present during the VLM pre-training phase, MDPR firstly introduces a multi-dimensional prompt construction strategy. During training, it leverages zero-shot VLMs to extract and construct a prompt pool for each class, capturing five distinct dimensions: general appearance, fine-grained appearance, functionality, contextual information, and differential features. This multi-faceted prompt design helps mitigate prior biases for classes. Subsequently, in the dynamic prompt routing stage, it further alleviates the impact of imbalanced data by implementing global visual-class alignment, dynamic routing-based visual-prompt matching, and fine-grained semantic balancing. This process generates predictions from multiple perspectives, and robust results are achieved through a logit fusion mechanism. As an effective enhancement architecture, the proposed MDPR can be flexibly integrated with various VLM fine-tuning methods. The codes are available in \url{https://anonymous.4open.science/r/MDPR-328C/README.md}.

To evaluate the effectiveness of the proposed MDPR, we conducted extensive experiments on three long-tailed visual recognition benchmarks, namely CIFAR-100-LT, ImageNet-LT, and Places-LT. The experimental results demonstrate that MDPR, through the comprehensive prompt construction and dynamic routing mechanisms, effectively mitigates class imbalance biases in both pre-trained models and downstream data, achieving robust performance improvements across head and tail classes while maintaining high compatibility with existing fine-tuning frameworks. The primary contributions of this work are:

\begin{itemize}
    \item We propose a plug-and-play framework for VLM's fine-tuning, termed MDPR, which addresses the challenge of joint imbalance through dynamic prompt routing, achieving efficient performance enhancement.
    \item We propose a multi-dimensional prompt construction approach, which systematically enhances the semantic understanding of VLMs by integrating five semantic dimensions, significantly mitigating inherent biases in pre-trained models.
    \item We validate the versatility of MDPR with different tuning methods, and it improves performance across three benchmarks with minimal additional parameters or time, particularly enhancing recognition of tail classes.
\end{itemize}
% 
% \begin{itemize}
%     \item . % 我们设计了基于语义-视觉表征研究和
%     \item . % 
%     \item . % 实验
% \end{itemize}
\section{Related Works}
\label{sec:related_works}

\subsection{Pretrained Model Fine-tuning under Long-tailed Distribution}
\label{sec:rw_long_tail_ft}
Long-tailed data distributions challenge pretrained model fine-tuning, often leading to a bias towards head classes and impairing generalization to tail classes. Traditional strategies such as re-balancing~\cite{shi2024efficient,tan2020equalization,cui2019class}, information augmentation~\cite{xu2023learning,li2024role}, and Mixture-of-Experts (MoE) models~\cite{fedus2022switch,zhang2023empowering} offer foundational solutions. More recently, novel fine-tuning approaches for multimodal pretrained models have been explored. \textbf{Cross-modal collaborative fine-tuning} enhances minority class representations via visual-semantic contrastive learning and feature alignment~\cite{chen2024multimodal}. \textbf{Parameter-efficient fine-tuning (PEFT)} techniques, including adapter tuning~\cite{kim2024longtailed} and prompt tuning~\cite{dong2022lpt}, aim to adjust for minority classes with minimal backbone alteration, mitigating overfitting. Furthermore, \textbf{knowledge transfer and distillation} leverage priors from large pretrained models, employing teacher-student paradigms or cross-domain transfer to bolster tail class robustness~\cite{rangwani2024deitlt}.While these fine-tuning strategies address long-tailed distributions from various angles, many focus on re-weighting samples/losses or adapting model parameters. In contrast, MDPR introduces an explicit, structured semantic knowledge base and a dynamic routing mechanism, offering a complementary pathway to directly enhance the semantic understanding and discriminative capability for classes, especially those in the tail.
\subsection{LLM-Enhanced Visual Representation Learning with Limited Samples}
\label{sec:rw_llm_enhanced}
Large Language Models (LLMs) have enriched visual learning in data-scarce scenarios like few-shot and long-tailed recognition. Research primarily explores three directions:
\textbf{Category semantic enhancement.} For fine-grained or underspecified labels, LLaMP~\cite{zheng2024large} employs LLMs to generate descriptive prompts, improving inter-class separability. ArGue~\cite{tian2024argue} integrates visual attributes and common sense semantics to guide prompt refinement. These methods typically yield a single, albeit enhanced, textual representation per class. MDPR, however, constructs a multi-dimensional prompt pool for each class, capturing diverse semantic facets, and dynamically selects from this pool based on image context, offering greater representational richness and adaptability.
\textbf{Sample generation.} In imbalanced settings, LLMs produce detailed descriptions to steer text-to-image (T2I) models for synthesizing tail-class samples, as in LTGC~\cite{zhao2024ltgc}. While effective for data augmentation, such approaches often incur significant computational overhead from generative models and may not directly enhance the VLM's intrinsic understanding. MDPR, instead, focuses on efficiently enriching the VLM with pre-computed semantic knowledge, rather than relying on external sample generation.
\textbf{Concept expansion.} LLMs facilitate modeling novel concepts in open-world settings. PerVL~\cite{cohen2022my} uses LLMs to generate personalized descriptions, extending VLM vocabularies. These methods primarily target open-set generalization or T2I generation. MDPR, while also leveraging LLM-derived knowledge, is specifically designed as a plug-and-play module to improve fine-tuning performance on closed-set, long-tailed recognition tasks by dynamically routing pre-defined, multi-faceted class semantics.
\section{Method}
\begin{figure*}[t]
  \includegraphics[width=\textwidth]{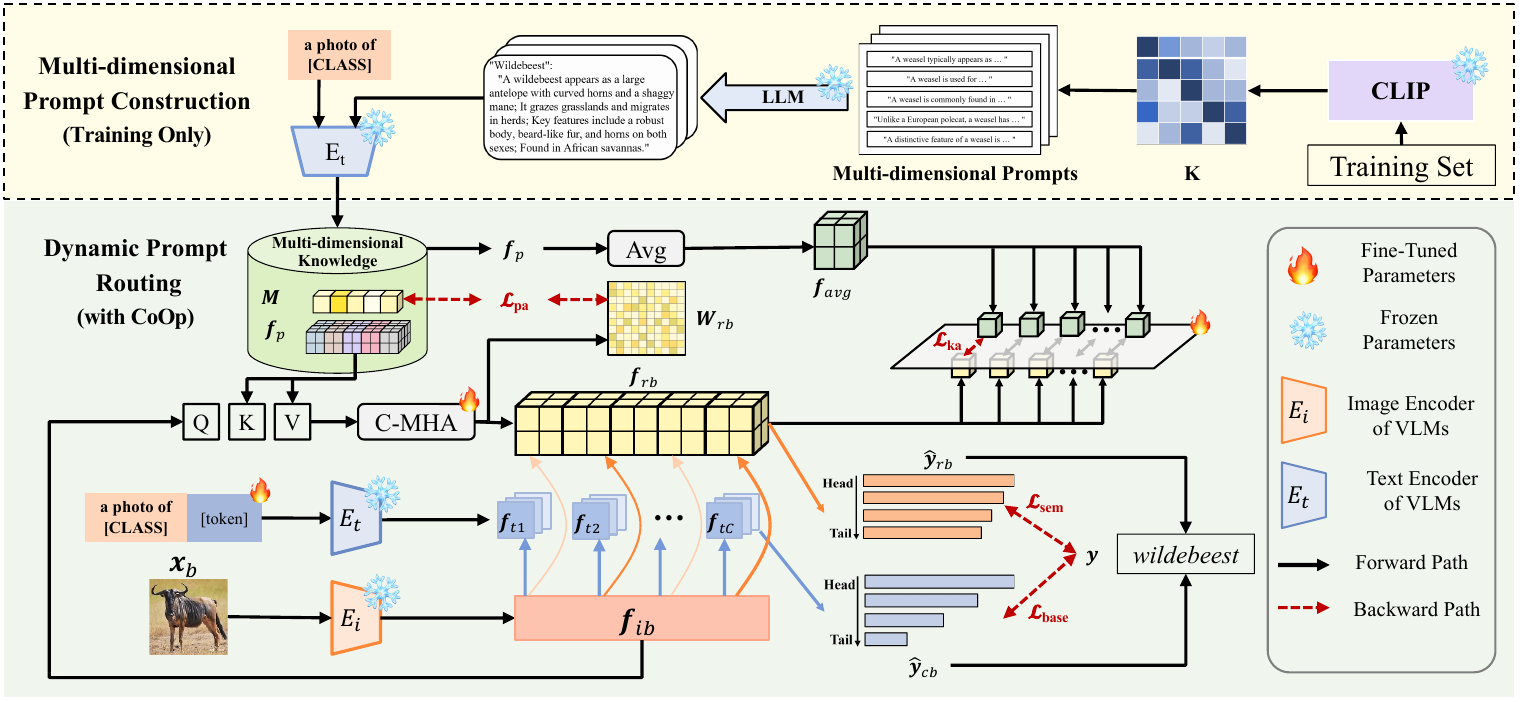}
  \caption{The framework of MDPR, which consists of two stages. The offline Multi-dimensional Prompt Construction builds a knowledge-base for enhancing semantics. The online Dynamic Prompt Routing aggregation the pre-learned knowledge for debias the predictions. Here we use CoOp \cite{zhou2022learning} for an example.}
  \label{fig:framework}
  \vspace{-0.5cm}
\end{figure*}
\subsection{Overview}
To alleviate biases in the fine-tuning of VLMs, the proposed Multi-dimensional Dynamic Prompt Routing (MDPR) routes a visual-semantic knowledge base to enhance representation learning. Figure \ref{fig:framework} shows the framework of MDPR, which comprises two synergistic modules: 1) The multi-dimensional prompt construction module generates a comprehensive knowledge base by designing five-dimensional prompts for LLM. 2) The dynamic prompt routing module enhances the utilization of the knowledge and provides de-biasing predictions. MDPR could serve as a flexible plug-and-play module, capable of seamless integration into existing fine-tuning methods.

\subsection{Multi-dimensional Prompt Construction}
To address potential inherent class biases in pre-trained VLMs, MDPR designs a class-specific prompt knowledge base spanning multiple semantic dimensions. The knowledge base endows VLMs with deeper understanding of classes, especially  for distinguishing similar classes or prior tail classes in pre-training data.

\subsubsection{Visual-Language Prompt Design}

Since a single class name often fails to capture the complex visual attributes of certain class, especially rare or nuanced concepts in real-world data. Recent works \cite{tan2024compound,zheng2024large} have explored using prompts that incorporate knowledge from LLMs or word knowledge to enhance visual representation learning. Based on this line of research, we conduct a review of existing prompting strategies and make a five-dimensional classification of prompts.

Moreover, considering the prior bias inherent in VLMs, we introduce an additional prompt dimension termed differential features. Given a dataset with $C$ classes, we first construct a confusion matrix $\mathbf{K}$ using CLIP's zero-shot predictions on the training set. For each class, the most frequently confused class is selected as the target for generating differential features, which may reflect biases in the model's pretraining distribution. To this end, the knowledge we includes following dimensions: 

\noindent
\textbf{General Appearance (GA)}: Typical visual features of the class, such as color, shape, and size \cite{tian2022vl,tan2024compound}.

\noindent
\textbf{Fine-grained Appearance (FA)}: Focuses on more specific local details and textures crucial for distinguishing sub-class or similar objects \cite{zheng2024large,tan2024compound,zhao2024ltgc}.

\noindent
\textbf{Functionality (FT)}: Articulates the primary function, purpose, or role of the class object in specific activities \cite{tian2024argue}.

\noindent
\textbf{Contextual Information (CI)}: Depicts the common background environments, associated objects, or typical scenarios where the class object is usually found \cite{tian2024argue,zhao2024ltgc}. 

\noindent
\textbf{Differential Features (DF)}: Highlights unique characteristics by contrasting the class with one or more easily confusable similar classes. 

For each class $c$, we obtain a set of prompts $\mathcal{P}_c = \{\text{p}_{c,v}\}_{v=1}^{V_{dim}}$, covering $V_{dim}$ prompts. The detailed prompt generation procedure and the choice of LLM are elaborated in Appendix \ref{appendix:llm_selection_brief}.

\subsubsection{Knowledge Base Construction}
The generated prompts are diverse from classes to dimensions, we then send the prompts to VLMs for organizing a class-specific knowledge base. The text encoder $E_{t}(\cdot)$ of a frozen CLIP encodes $\text{p}_{c,v}$ for $v$-th dimension prompt of class $c$ into a $d$-dimensional feature $\mathbf{f}^{cv}_{p}= E_{t}(\text{p}_{c,v})$, where $d$ is the latent dimension of CLIP.

The encoded features combine as multi-dimensional prompt features $\mathbf{f}_{p} \in \mathbb{R}^{C \times V_{dim} \times d}$. To obtain a more general class-level semantic representation, the $V_{dim}$ dimensional prompt features $\mathbf{f}^{cv}_{p}$ are averaged to $\mathbf{f}^{c}_{avg}\in\mathbb{R}^d$:
\begin{equation}
\label{eq:avg_sem_feat_detail}
\mathbf{f}^{c}_{avg} = \frac{1}{V_{dim}} \sum_{v=1}^{V_{dim}} \mathbf{f}^{cv}_{p}
\end{equation}

To introduce a beneficial inductive bias during dynamic routing and to guide the learning of attention weights, we construct a prior alignment matrix $\mathbf{M} \in \mathbb{R}^{C \times V_{dim}}$ . An element $\mathbf{M}[c,v]$ represents the prior importance of the $v$-th dimensional prompt for class $c$. This importance is defined by the similarity between the encoded $v$-th dimensional prompt $\mathbf{f}_{p}^{c,v}$ and the encoding of a standard generic $prompt$ for that class (e.g., "a photo of a [class name $c$]"):
\begin{equation}
\label{eq:prior_matrix_m_detail}
\mathbf{M}[c,v] = \operatorname{Sim}(\mathbf{f}_{p}^{c,v}, E_{t}(prompt(c)))
\end{equation}

% The matrix $\mathbf{M}$ will be utilized as a target in the prior alignment loss ($\mathcal{L}_{\text{upd}}$) to regularize the dynamically learned attention weights, guiding the model to focus on dimensions more relevant to the core semantics of the class.

\subsection{Dynamic Prompt Routing}
The Dynamic Prompt Routing (DPR) module designs relevant semantic information from the pre-constructed multi-dimensional prompts for each class, conditioned on the visual context of the input image. This process generates fine-grained, class-aware semantic representations for tail classes.

\subsubsection{Image-attentive Semantic Extraction}
\label{sec:dynamic_semantic_generation}

The aim of this stage is to learn an instance-aware, dynamically adjusted semantic representation for visual inputs. In a long-tailed visual recognition dataset $D=\{X,Y\}$, for an image $x_{b}\in X$ labeled $y_{b}$, its $d$-dimensional visual features $\mathbf{f}_{ib} = E_{i}(x_{b})$ are extracted by the image encoder $E_{i}(\cdot)$ of pre-trained CLIP.

The learned $\mathbf{f}^{b}_{i}$ underlines certain aspects of the semantics of class, a class-specific multi-head attention (C-MHA) module computes attention weights and forms attentive semantic features. For an image $x_{b}$ in class $c$, the C-MHA outputs $\mathbf{W}^{c}_{r}$ for routing and attentive semantic features $\mathbf{f}^{c}_{rb}$, which can be formulated as:
% over the $V_{dim}$ semantic dimensions for class $c$. These weights signify the importance of each semantic dimension of class $c$ in the context of the current image $b$. 
% Subsequently, these attention weights are used to perform a weighted aggregation of the $V_{dim}$ prompt features of class $c$ (from $\mathbf{F}_{\text{prompt}}[c,:,:] \in \mathbb{R}^{V_{dim} \times d}$), yielding a dynamic semantic representation $\tilde{\mathbf{x}}_{\text{sem}}^{(b,c)} \in \mathbb{R}^d$ tailored for image $b$ and class $c$:

\begin{equation}
\label{eq:mha_dpr}
\mathbf{f}^{c}_{rb}, \mathbf{W}^{c}_{r} = \operatorname{C-MHA}(\mathbf{f}_{ib},\mathbf{f}^{c}_{p},,\mathbf{f}^{c}_{p})
\end{equation}

This C-MHA allows the model to dynamically focus on the most discriminative semantic aspects for each class conditioned on the image. In practice, we accelerate this procedure by a matrix manner, and collect the weights $\mathbf{W}_{r}$ and $\mathbf{f}_{rb}$ for image $x_{b}$ across classes.

\subsubsection{Semantic-enhanced Class Prediction}
\label{sec:dynamic_classification}

Leveraging the image features $\mathbf{f}_{ib}$ and the attentive semantic representations $\mathbf{f}_{rb}$, we formally compute semantic logits $\mathbf{\hat{y}}_{\text{rb}}$ for classification. For an image $b$ and class $c$, the $\mathbf{\hat{y}}_{\text{rb}}$ is defined as:
\begin{equation}
\label{eq:dynamic_logits}
\mathbf{\hat{y}}_{\text{rb}} = s \cdot \langle \mathbf{f}_{rb},\mathbf{f}_{ib} \rangle
\end{equation}
where $s$ is a learnable temperature parameter. The logits $\mathbf{\hat{y}}_{\text{rb}}$ directly reflect the model's confidence in assigning an image to each class based on the dynamically aggregated semantic information.

Correspondingly, we introduce a dynamic semantic loss ($\mathcal{L}_{\text{sem}}$) to supervise this classification branch. This loss holds comparable importance to the base VLM's classification loss ($\mathcal{L}_{\text{ce}}$) and jointly drives the primary classification task:
\begin{equation}
\label{eq:loss_sem}
\mathcal{L}_{\text{sem}} = \text{CLA}(\mathbf{z}_{\text{sem}}, \mathbf{y})
\end{equation}
where $\mathbf{y}$ represents the ground-truth labels, and Compensated Cross Entropy refers to loss function designed for imbalanced data \cite{shi2024efficient}.

To resist the bias accumulation under long-tailed scene, we introduce a regularization loss $\mathcal{L}_{\text{reg}}=\lambda_{\text{pa}}\mathcal{L}_{\text{pa}}+\lambda_{\text{ka}}\mathcal{L}_{\text{ka}}$, where $\lambda_{\text{pa}}$ and $\lambda_{\text{ka}}$ are weights of losses. The $\mathcal{L}_{\text{pa}}$ and $\mathcal{L}_{\text{ka}}$ target the attention routing strategy and the quality of the generated dynamic semantic representations, respectively:

\paragraph{Prior Alignment Loss for Routing Strategy Optimization ($\mathcal{L}_{\text{pa}}$):}
This loss aims to guide the learning of weights $\mathbf{W}_{r}$ using the prior alignment matrix $\mathbf{M}$, and $\mathbf{M}[c,v]$ signifies the prior importance of the $v$-th semantic dimension for class $c$. 

It encourages the learned attention distribution to align with this desirable prior distribution, which may be more balanced or incorporate domain knowledge, thereby optimizing the information routing strategy:
\begin{equation}
\label{eq:loss_upd}
\mathcal{L}_{\text{pa}} = \frac{1}{C} \sum_{c=1}^{C} \left(1 - \text{Sim}(\mathbf{w}_{\text{r}}^{(c)}, \mathbf{M}[c,:])\right)
\end{equation}

This alignment helps the model to focus on semantically crucial dimensions, mitigating the influence of statistical biases in the data during dynamic routing.

\paragraph{Knowledge Alignment Loss for Representation Quality Enhancement ($\mathcal{L}_{\text{ka}}$):}
To improve the stability and generalization of the instance-aware dynamic semantic features $\mathbf{f}_{rb}$ via knowledge distillation. DPR encourages the distribution of the dynamic semantic features for the ground-truth class $y_b$ (after a learnable linear projection $\text{Proj}(\cdot)$) to align with the distribution of the averaged semantic features for that class $\mathbf{f}^{y_b}_{avg}$ (i.e., $\mathbf{f}^{y_b}_{avg}[y_b,:]$, passed through the same projection layer) in a high-dimensional space:
\begin{equation}
\label{eq:loss_kl}
\mathcal{L}_{\text{ka}} = \operatorname{KL}\left(\operatorname{Proj}(\mathbf{f}_{rb},\operatorname{Proj}(\mathbf{f}^{y_b}_{avg}[y_b,:]))\right)
\end{equation}
where ${\text{KL}(\cdot,\cdot)}$ denotes the KL divergence.

% \subsubsection{Prompt Construction}
% \subsubsection{Training}
\subsection{Training Strategy}

The training objective of the MDPR framework is to optimize the entire model end-to-end, enabling it to effectively leverage the multi-dimensional semantic knowledge base through dynamic routing for superior performance on imbalanced downstream visual tasks. The multi-task loss synergistically optimizes the representational capacity of the base VLM and the semantic enhancement and regularization mechanisms of the MDPR module.

\subsubsection{Learnable Parameters}
During the training process, the following parameters are subject to optimization:

\noindent
\textbf{Base VLM Framework Parameters:} Depending on the chosen base VLM fine-tuning paradigm (e.g., CoOp or MaPLe), this may include its learnable prompt parameters (e.g., CoOp's context vectors $\mathbf{ctx}$, MaPLe's multi-level prompt parameters).

\noindent
\textbf{MDPR Module Parameters:} This encompasses the parameters of the C-MHA, and the parameters of the learnable linear projection layer $\text{Proj}(\cdot)$.

The pre-computed multi-dimensional prompt features $\mathbf{f}_{p}$, class-level averaged semantic features $\mathbf{f}_{avg}$, and the prior alignment matrix $\mathbf{M}$ serve as fixed inputs during the training phase and are not updated.

\subsubsection{Optimization}

The MDPR model is optimized by minimizing the sum of losses. The total loss function $\mathcal{L}_{\text{total}}$ is formulated as:
\begin{equation}
\label{eq:final_total_loss}
\mathcal{L}_{\text{total}} = \lambda_{\text{base}}\mathcal{L}_{\text{base}} + \lambda_{\text{sem}}\mathcal{L}_{\text{sem}} + \mathcal{L}_{\text{reg}}
\end{equation}
where $\lambda_{\text{base}}$ and $\lambda_{\text{sem}}$ are the weights of losses, and $\lambda_{\text{base}}$ is typically set to 1.0.

\subsection{Logits-fused Inference}
\label{sec:inference_strategy}
During inference, MDPR combines the predictions from the base VLM framework and the dynamic semantic routing pathway to yield final predictions. Given the class name-based logits $\hat{\bf y}_{cb}$ and the routing-based logits $\hat{\bf y}_{rb}$, the final fused logits $\hat{\bf y}_{fuse}$ are computed as:
\begin{equation}
\label{eq:logit_fusion}
\hat{\bf y}_{fuse} = (1-\beta) \cdot \hat{\bf y}_{cb} + \beta \cdot \hat{\bf y}_{rb}
\end{equation}
where $\beta \in [0, 1]$ is a hyperparameter balancing the two sources, typically set to $0.5$ in our experiments (Section~\ref{sec:impl_details_sub}). This fusion allows MDPR to benefit from both the general representations of the base VLM and the instance-specific insights from the DPR module.

\section{Experiments}
\label{sec:experiments_main}

To comprehensively evaluate the efficacy of MDPR, we conduct extensive experiments on three long-tailed image recognition benchmarks. 
% We first introduce the datasets employed and detail our experimental setup, including the evaluation metrics. 
% Subsequently, we will compare MDPR (integrated with CoOp and MaPLe frameworks, denoted as MDPR-CoOp and MDPR-MaPLe, respectively) against representative state-of-the-art baseline methods. Finally, through thorough ablation studies and analyses, we will investigate the contributions of MDPR's key components and their impact on model performance.

\subsection{Datasets}
\label{sec:datasets_exp_detailed}

Our experiments are conducted on three widely adopted long-tailed image recognition benchmarks: CIFAR-100-LT \cite{cao2019learning}, ImageNet-LT \cite{liu2019large}, and Places-LT \cite{liu2019large}. Detailed statistics for these datasets are presented in Table~\ref{tab:dataset}.
\begin{table}

  \centering
  \small
  \begin{tabular}{ccccc}
    \hline
    \textbf{Dataset} & \textbf{\#Class} & \textbf{IR} & \textbf{\#Train}& \textbf{\#Test}\\
    \hline
    \multirow{3}{*}{CIFAR-100-LT} & \multirow{3}{*}{100} & {10} & {19,573} & \multirow{3}{*}{10,000} \\
    {} & {} & {50} & {12,608} & {} \\
    {} & {} & {100} & {10,847} & {} \\
    {ImageNet-LT} & {1,000} & {256} & {115,846} & {50,000} \\
    {Places-LT} & {365} & {996} & {62,500} & {7,300} \\
    \hline
  \end{tabular}
  \caption{\label{tab:dataset} Statistics of long-tailed datasets, where "\#" means the number of item.}
  \label{dataset_stats_exp_detailed}
\end{table}

\subsection{Experimental Settings}

\subsubsection{Evaluation Metrics}
Following the evaluation protocol proposed in~\cite{liu2019large}, we report accuracies of all classes and three class subsets: Many-classes (>100 images), Medium-classes (20-100 images), and Few-classes (<20 images). This detailed breakdown allows for a more nuanced understanding of model behavior across varying class data densities.

\subsubsection{Implementation Details}
\label{sec:impl_details_sub}

\textbf{Base VLM Framework and Backbone:}
The MDPR is implemented and evaluated on top of two prominent prompt learning frameworks: CoOp \cite{zhou2022learning} and MaPLe \cite{khattak2023maple}. These are referred to as MDPR-CoOp/Ours(CoOp) and MDPR-MaPLe/Ours(MaPLe), respectively. All experiments except the CPRL \cite{yan2024category} utilize the pre-trained CLIP ViT-B/16 model as the visual backbone.

\noindent
\textbf{Training Hyperparameters:}
All models are trained using the AdamW optimizer with a weight decay of $1 \times 10^{-4}$. The initial learning rate is set to $1 \times 10^{-3}$, decayed using a cosine annealing schedule over $20$ epochs. A batch size of $128$ is used for all datasets. The loss weights $\lambda_{\text{base}}, \lambda_{\text{sem}}, \lambda_{\text{pa}}, \lambda_{\text{ka}}$ in Equation (\ref{eq:final_total_loss}) are determined through systematic tuning, with $\lambda_{\text{base}}$ fixed at $1.0$. The weights for $\mathcal{L}_{\text{sem}}$ and $\mathcal{L}_{\text{ka}}$ are linearly warmed up from $0$ to their target values over the first $5$ epochs. The logit fusion coefficient $\beta$ (for combining $\mathbf{z}_{\text{base}}$ and $\mathbf{z}_{\text{sem}}$ during inference, see Equation~\ref{app:inference_settings} is set to $0.5$ by default. All experiments were conducted on a single NVIDIA RTX 3090 GPU. Further details on hyperparameter tuning ranges, final selected values, and the KL temperature $T$ are provided in Appendix~\ref{app:hyperparams}. % Adjust appendix label as needed

\begin{table*}[htbp]
  \centering
  \small
  \begin{tabular}{l|lll|llll|llll}
    \hline
    \multirow{2}{*}{\bf Model} & \multicolumn{3}{c}{\bf IR=10} & \multicolumn{4}{c}{\bf IR=50} & \multicolumn{4}{c}{\bf IR=100} \\
    \cline{2-12}
    {} & {\bf All} & {\bf Many} & {\bf Med} & {\bf All} & {\bf Many} & {\bf Med} & {\bf Few} & {\bf All} & {\bf Many} & {\bf Med} & {\bf Few} \\
    \hline
    CLIP-ViT-B/16 (ICML'21) & 59.50 & 61.09 & 55.97 & 59.50 & 64.05 & 57.27 & 54.22 & 59.50 & 61.83 & 59.74 & 56.50 \\
    \hline
    CoOp (IJCV'22) & 70.88 & 75.06 & 61.58 & 65.70 & 79.63 & 58.44 & 50.50 & 64.34 & 79.43 & 64.51 & 46.53 \\
    CoCoOp (CVPR'22) & 72.29 & 76.75 & 62.35 & 66.38 & 80.20 & 60.20 & 49.00 & 63.90 & 80.69 & 65.20 & 42.80 \\
    MaPle (CVPR'23) & \underline{81.98} & 84.58 & \underline{76.19} & \underline{77.09} & \underline{87.34} & \underline{71.98} & 65.39 & \underline{74.09} & \underline{88.14} & \underline{73.46} & 58.43 \\
    MaPle+LA(NeurIPS 2020)  & 78.26 & 80.93 & 72.32 & 75.01 & 83.02 & 69.37 & 69.61 & 72.05 & 84.00 & 70.69 & 59.70 \\
    PLOT++ (ICLR'23) & 75.52 & 78.83 & 68.16 & 70.73 & 82.95 & 64.54 & 57.00 & 68.37 & 81.89 & 67.54 & 53.57 \\
    LASP (CVPR'23) & 68.57 & 72.64 & 59.52 & 63.76 & 76.49 & 57.15 & 49.83 & 61.29 & 76.49 & 60.17 & 44.87 \\
    TextRefiner (AAAI'25) & {74.22} & {78.12} & {65.55} & {67.70} & {81.83} & {62.51} &{47.33} & {64.32} & {83.00} & {66.03} & {40.53}\\
    CPRL (MM'24) & 81.75 & \underline{84.97} & 74.58 & 71.16 & 86.61 & 65.22 & 49.50 & 68.20 & {\bf 88.74} & 71.97 & 39.83 \\
    Candle (KDD'24) & 75.77 & 76.71 & 73.68 & 73.14 & 77.15 & 70.17 & \underline{70.78} & 72.42 & 76.14 & 72.54 & {\bf 67.93} \\
    \hline
    Ours (CoOp) & {76.25} & {78.51} & {71.23} & {72.44} & {81.76} &  {67.93} & {61.50} & {70.33} & {81.17} & {70.57} & {57.40} \\
    Ours (Maple) & {\bf 84.73} & {\bf 86.32} & {\bf 81.19} & {\bf 81.38} & {\bf 88.68} &  {\bf 76.90} & {\bf 74.94} & {\bf 79.25} & {87.60} & {\bf 81.26} & {\underline{67.17}} \\
    \hline
  \end{tabular}
    \caption{Comparison results on CIFAR-100-LT dataset, where best results are {\bf bolded} and suboptimal results are \underline{underlined}. According to the split standard of dataset, CIFAR-100-LT with IR=10 contains no few-sampled classes.}
    \label{tab:performance1}
\end{table*}

\subsection{Comparison Results}
\label{sec:main_comparison_results_moderately_condensed_formatted}
To comprehensively evaluate the MDPR framework's effectiveness in addressing long-tailed distributions, this section presents a comparative performance analysis against a range of representative methods on CIFAR-100-LT, ImageNet-LT, and Places-LT. The compared methods include Zero-Shot CLIP (ZS CLIP)~\cite{radford2021learning}, mainstream VLM prompt tuning approaches such as CoOp~\cite{zhou2022learning}, CoCoOp~\cite{zhou2022conditional}, MaPLe~\cite{khattak2023maple}, LASP~\cite{bulat2023lasp}, and PLOT~\cite{chen2022plot}, recent VLM enhancement (e.g., TextRefiner~\cite{xie2025textrefiner}), long-tail recognition techniques (e.g., CPRL~\cite{yan2024category}, Candle\cite{shi2024efficient}, and our reproduced MaPLe+LA~\cite{ren2020balanced} baseline). All experiments were conducted under fair conditions. MDPR integrated with CoOp and MaPLe is denoted as \textbf{Ours (CoOp)} and \textbf{Ours (MaPLe)}. Detailed classification accuracies are in Tables~\ref{tab:performance1} and~\ref{tab:performance2}
% Firstly, standard VLM prompt tuning methods generally exhibit significant head-class bias under long-tailed distributions, leading to notably degraded tail-class recognition, sometimes underperforming ZS CLIP. As detailed in Tables \ref{tab:performance1} and \ref{tab:performance2}, under high imbalance scenarios like CIFAR-100-LT (IR=100) and Places-LT, the Few-shot accuracies of methods like CoOp and MaPLe are considerably lower than ZS CLIP, highlighting their struggle to overcome biases from data sparsity with limited label supervision.

 \textbf{The proposed MDPR framework substantially enhances base VLM fine-tuning performance, achieving consistent and significant gains across all class groups and scenarios, particularly reaching SOTA levels for Few-shot classes on multiple benchmarks.} For instance, on the challenging CIFAR-100-LT (IR=100), Ours (CoOp) and Ours (MaPLe) improve Few-shot accuracy from CoOp's 46.53\% and MaPLe's 58.43\% to 57.40\% and 67.17\%, respectively. On larger datasets like ImageNet-LT and Places-LT, MDPR also demonstrates strong efficacy; notably, Ours (MaPLe) boosts Few-shot accuracy on Places-LT by over 22 p.p. compared to MaPLe, securing top performance in Overall, Medium-shot, and Few-shot metrics on several datasets. These results robustly validate that MDPR, via structured multi-dimensional semantics and image-conditioned dynamic routing, effectively supplements VLMs with discriminative information, enhancing representation learning for balanced performance on long-tailed data.
 
 \textbf{MDPR demonstrates universality and effectiveness as an enhancement module across different base frameworks and datasets.} While MaPLe inherently outperforms CoOp on some datasets, MDPR consistently delivers significant gains when combined with either framework. The substantial Few-shot improvement MDPR brings to MaPLe on Places-LT (over 22 p.p.) compared to that for CoOp (approx. 18 p.p.) suggests its particular effectiveness in unlocking the potential of advanced frameworks under extreme imbalance. Furthermore, unlike some specialized long-tail methods (e.g., Candle) that might excel on tail classes for specific datasets at the cost of head/medium class performance, MDPR promotes more balanced improvements.
 
 \textbf{MDPR's relative advantage tends to be more pronounced at higher imbalance ratios.} Comparing results on CIFAR-100-LT across increasing IRs shows that while all methods' absolute performance declines, MDPR's (especially Ours (MaPLe)) improvement margin over baselines often widens. This further substantiates the crucial role of MDPR's multi-dimensional semantic understanding and dynamic routing in tackling extreme data imbalance.
 
\begin{table*}[htbp]
  \centering
  \small
  \setlength{\tabcolsep}{0.3cm}{
  \begin{tabular}{l|llll|llll}
    \hline
    \multirow{2}{*}{\bf Model} & \multicolumn{4}{c}{\bf ImageNet-LT} & \multicolumn{4}{c}{\bf Places-LT}\\
    \cline{2-9}
    {} & {\bf All} & {\bf Many} & {\bf Med} & {\bf Few} & {\bf All} & {\bf Many} & {\bf Med} & {\bf Few} \\
    \hline
    CLIP-ViT-B/16 (ICML'21) & {62.95} & {63.96} & {62.08} & {63.15} & {38.40} & {35.49} & {37.97} & {44.77} \\
    \hline
    CoOp (IJCV'22) & {68.69} & {74.82} & {65.75} & {61.75} & {40.73} & \underline{53.10} & {35.58} & {29.72} \\
    CoCoOp (CVPR'22) & {-} & {-} & {-} & {-} & {41.12} & {52.95} & {35.84} & {31.39} \\
    MaPle (CVPR'23) & {69.02} & {77.20} & {64.90} & {60.37} & {41.37} & \textbf{54.33} & {36.45} & {28.73} \\
    MaPle+LA(NeurIPS 2020)  & 70.78 & 78.13 & 67.53 & 61.46 & 44.99 & 51.75 & 41.04 & 41.59  \\
    TextRefiner (AAAI'25) & {66.74} & \textbf{81.75} & {62.45} &{39.40} & {38.01} & {52.76} & {32.79} & {22.79}\\
    Candle (KDD'24) & {71.28} & {76.38} & {69.55} & {62.91} & {45.81} & {46.97} & {45.42} & {44.56} \\
    \hline
    Ours (CoOp) & \underline{74.57} & {77.67} &  \underline{73.42} & \underline{69.87} & \underline{48.89} & {49.45} &  \underline{48.72} & \underline{47.96}  \\
    Ours (Maple) &\textbf{75.57} & \underline{79.42} & \textbf{74.02} & \textbf{70.04}  & \textbf{50.94} & {50.32} &  \textbf{51.42} & \textbf{50.99} \\
    \hline
  \end{tabular}
  \caption{\label{tab:performance2} Comparison results on ImageNet-LT dataset and Places-LT dataset, where best results are {\bf bolded} and suboptimal results are \underline{underlined}. The "-" in results means out of memory in our devices.}
  }
\end{table*}
\begin{figure}[t]
  \includegraphics[width=\linewidth]{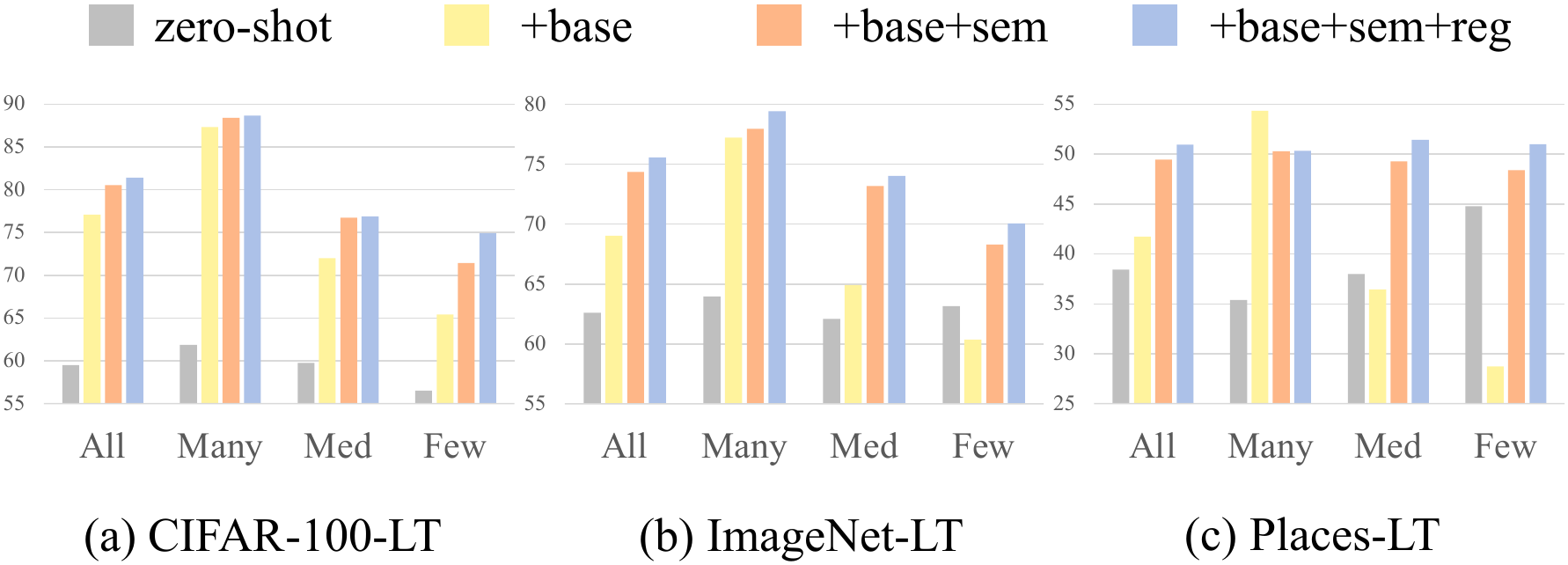}
  \caption{Ablation results across datasets. Results of other IR of CIFAR-100-LT see in Appendix \ref{app:ablation}.}
  \label{fig:ablation}
\end{figure}

\subsection{Ablation Study}
Our algorithm achieves performance gains through stacking multi-dimensional semantic prompts and regularization modules. To assess their contributions, we compared zero-shot CLIP, base MaPLe(\textbf{base}), MaPLe with semantic prompts (\textbf{base+Sem}), and full MDPR (\textbf{base+sem+reg}) on CIFAR-100-LT (IR=50), ImageNet-LT, and Places-LT. As shown in Figure~\ref{fig:ablation}, performance improves progressively with each module. Adding semantic prompts (\textbf{base+sem}) significantly boosts tail-class accuracy, e.g., Few on Places-LT from 28.73\% to 48.39\% (+19.66\%). Regularization (\textbf{base+sem+reg}) further raises Few to 50.99\% and slightly improves head and mid classes (e.g., Many to 79.42\% on ImageNet-LT). Semantic prompts substantially mitigate tail-class bias via comprehensive semantic representations, while regularization enhances prediction consistency across head, mid, and tail classes by stabilizing dynamic routing.
\subsection{In-depth Analysis of Knowledge-base Construction}
\label{sec:ablation_knowledge_base_condensed}

An ablation study on the multi-dimensional semantic knowledge base using \textbf{Ours (CoOp)} on Places-LT (Table~\ref{tab:ablation_results}) reveals each dimension's distinct contribution. Removing the Differential Features (DF) dimension caused the largest overall accuracy drop, highlighting the critical role of distinguishing unique characteristics via comparison with similar classes. The removal of Contextual Information (CI) or General Appearance (GA) also significantly impacted performance, underscoring the importance of scene understanding and fundamental visual features.In contrast, lacking Fine-grained Appearance (FA) or Functionality (FT) had a smaller, yet noticeable, negative effect, confirming the supplementary value of specific visual details and object function information. Notably, all dimensions positively contributed to few-shot class recognition; removing any single dimension decreased few-shot accuracy by 1.3\% to 2\%, with CI removal having the most pronounced effect.

These findings demonstrate that a comprehensive, multi-dimensional knowledge base with complementary semantic dimensions is essential for MDPR to effectively address long-tailed distributions and enhance learning of data-scarce classes.

\begin{table}[htbp]
\centering

\begin{tabular}{lcccc}
\hline{}
\textbf{Knowledge} & \textbf{All} & \textbf{Many} & \textbf{Mid} & \textbf{Few} \\
\hline{}
All & \textbf{48.89} & \textbf{49.45} & \textbf{48.72} & \textbf{48.24} \\
w/o GA &47.23 & 48.24 &46.70 &46.58 \\
w/o FA &47.88 &48.67 &47.53 &47.24 \\
w/o FT &47.87 &49.10 &47.04 &47.48 \\
w/o CI &47.28 &49.40 &46.05 &46.21 \\
w/o DF &46.82 &48.11 &45.74 &46.92 \\
\hline{}
\end{tabular}
\caption{ Different knowledge base on Places-LT.}
\label{tab:ablation_results}
\vspace{-0.2cm}
\end{table}

\subsection{Comparative Analysis of Model Efficiency}
\label{sec:param_efficiency_final}
To assess the practical applicability of our proposed MDPR framework, this section briefly analyzes the additional parameter count and its impact on training efficiency.
As summarized in Table~\ref{tab:param_efficiency_final}, our MDPR module introduces approximately 1.1M trainable parameters.
This increment is substantially smaller than the total parameter count of the CLIP ViT-B/16 backbone (representing less than 0.74\% of the backbone's parameters), positioning MDPR within the realm of parameter-efficient fine-tuning.
For training on the ImageNet-LT dataset, integrating MDPR results in a slight increase in per-epoch training time of approximately 14 seconds for the CoOp baseline and 114 seconds for the MaPLe baseline.

In summary, while MDPR introduces a modest number of additional parameters and a slight increase in computation, these are well-justified by the significant performance gains, particularly in recognizing few-shot classes.
The marginal overhead is especially low when MDPR is integrated with more complex frameworks like MaPLe, underscoring its practicality as an efficient enhancement module for VLMs addressing imbalanced data.

\begin{table}[htbp!]
  \centering
  \small
    
  \begin{tabular}{lcccc}
    \hline
    \multirow{2}{*}{\textbf{Metric}}&\multirow{2}{*}{\textbf{CoOp}}&\textbf{CoOp} & \multirow{2}{*}{\textbf{MaPLe}}&\textbf{MaPLe} \\
    {} & {} & \textbf{+MDPR} & {} & \textbf{+MDPR} \\
    \hline
    Param (M) & 0.008 & 1.108 & 3.6 & 4.7 \\
    
    Time (s) & 1115  & 1229   & 1361  & 1375  \\
    \hline
  \end{tabular}
  \caption{Trainable Parameters (Param) and Training Time per Epoch (Time) on ImageNet-LT.}
  \label{tab:param_efficiency_final}

\end{table}

\section{Conclusion}
\label{sec:bibtex}
Addressing the class bias in fine-tuning vision-language models under long-tailed distributions, we propose the Multi-dimensional Dynamic Prompt Routing (MDPR) framework. Unlike traditional static prompt or high-cost sample generation methods, MDPR leverages a structured multi-dimensional semantic knowledge base and an image-driven dynamic routing mechanism to efficiently mitigate biases from pre-training and downstream data. First, MDPR constructs a five-dimensional prompt pool, providing comprehensive class understanding to counter prior biases. Second, an image-guided dynamic routing module, combined with regularization, generates instance-adaptive class representations by optimizing routing and representation stability. Experiments on CIFAR-100-LT, ImageNet-LT, and Places-LT demonstrate that MDPR significantly enhances tail-class performance while balancing head and medium-class robustness, achieving SOTA or highly competitive results. As a lightweight plug-and-play module, MDPR offers an effective paradigm for open-world long-tailed recognition.

\section*{Limitations}
\begin{itemize}
    \item Limited by the devices, the effectiveness of MDPR has been primarily validated on the CLIP ViT-B/16 backbone integrated with CoOp and MaPLe. Its generalizability and performance on larger-scale or different VLM architectures require further examination in future work.
    \item MDPR's prediction balancing, while benefiting from the rich multi-dimensional semantic library, still partially relies on known class distribution information from the training set. This dependency might limit its robustness in real-world scenarios with unknown or dynamic class distributions. Future research could explore integrating methods like causal inference to enhance adaptability to open environments.
    \item The current multi-dimensional semantic knowledge base is constructed offline for predefined classes, potentially posing scalability challenges in incremental or open-set learning scenarios requiring rapid adaptation to new classes or domains. Future work could draw from continual learning principles to explore mechanisms for dynamic construction and updating of the semantic knowledge base.
\end{itemize}

\bibliography{acl_latex}

\appendix
\appendix
\section{Appendix} % Or just \appendix if your template handles sectioning

% You might have other appendix sections, like Detailed Comparison Results if they were moved here,
% or Prompt Generation Details. This example focuses on Hyperparameters.

\subsection{Detailed Hyperparameter Settings}
\label{app:hyperparams} % Label for this appendix subsection

This section provides a comprehensive overview of the hyperparameter settings used for training our MDPR models and the baseline VLM frameworks, supplementing the details in Section~\ref{sec:impl_details_sub} of the main paper. All experiments were conducted on a single NVIDIA RTX 3090 GPU.

\subsubsection{Common Training Settings}
The following settings were applied to all trained models (both baselines and our MDPR variants) unless specified otherwise:
\begin{itemize}
    \item \textbf{Optimizer:} AdamW~\cite{loshchilov2017decoupled}.
    \item \textbf{Weight Decay:} $1 \times 10^{-4}$.
    \item \textbf{Base Learning Rate (for prompts and MDPR modules):} $1 \times 10^{-3}$.
    \item \textbf{Learning Rate Schedule:} Cosine annealing schedule.
    \item \textbf{Total Training Epochs:} $20$.
    \item \textbf{Batch Size:} $128$ for all datasets.
    \item \textbf{Visual Backbone:} Pre-trained CLIP ViT-B/16~\cite{radford2021learning} for all experiments. The backbone parameters were kept frozen during fine-tuning of CoOp, MaPLe, and their MDPR-enhanced counterparts, consistent with standard prompt tuning practices.
\end{itemize}

\subsubsection{Base VLM Framework Parameters}
\label{app:base_vlm_params}
When MDPR is integrated, the parameters of the underlying base VLM frameworks were set as follows:

\noindent\textbf{CoOp~\cite{zhou2022learning}:}
\begin{itemize}
    \item \textbf{Number of Context Tokens ($N_{ctx}$):} 16.
    \item \textbf{Class Token Position:} ``end''.
    \item \textbf{Context Initialization:} Random initialization (standard for CoOp).
\end{itemize}

\noindent\textbf{MaPLe~\cite{khattak2023maple}:}
\begin{itemize}
    \item \textbf{Number of Context Tokens ($N_{ctx}$):} 2 for both visual and language shallow prompts.
    \item \textbf{Deep Prompt Depth (Vision \& Language):} 9 layers for both vision and language encoders.
    \item \textbf{Context Initialization:} Random initialization.
\end{itemize}

\subsubsection{MDPR Module Parameters}
\label{app:mdpr_module_params}
The specific parameters for our MDPR module were configured as:
\begin{itemize}
    \item \textbf{Semantic Prompt Embedding Dimension ($d$):} 512 (consistent with CLIP ViT-B/16 text encoder output).
    \item \textbf{Multi-Head Attention (MHA) in DPR:}
        \begin{itemize}
            \item Number of Attention Heads: 8.
            \item Dropout Rate (during training): 0.1.
        \end{itemize}
    \item \textbf{KL Projection Layer ($\text{Proj}(\cdot)$):} This linear layer projects features from $d=512$ to an intermediate dimension of $128$ before KL divergence calculation.
\end{itemize}

\subsubsection{Loss Weights and Temperatures}
\label{app:loss_weights_temps}
The weights for the individual loss components in the total loss function (Equation~\ref{eq:final_total_loss} in the main paper: $\mathcal{L}_{\text{total}} = \lambda_{\text{base}}\mathcal{L}_{\text{base}} + \lambda_{\text{sem}}\mathcal{L}_{\text{sem}} + \lambda_{\text{pa}}\mathcal{L}_{\text{pa}} + \lambda_{\text{ka}}\mathcal{L}_{\text{ka}}$) and the KL distillation temperature $T$ were determined through systematic tuning on a validation split (e.g., a subset of the training data or a dedicated validation set if available).

\begin{itemize}
    \item \textbf{$\lambda_{\text{base}}$:} Fixed at $1.0$ for all experiments to give primary importance to the base VLM's objective.
    \item \textbf{Tuning Strategy for $\lambda_{\text{sem}}, \lambda_{\text{pa}}, \lambda_{\text{ka}}, T$:}
        A two-stage tuning process was generally followed:
        \begin{enumerate}
            \item \textbf{Stage 1 (Tuning $\lambda_{\text{sem}}$):} $\lambda_{\text{pa}}$ and $\lambda_{\text{ka}}$ were initially set to $0$. $\lambda_{\text{sem}}$ was tuned by exploring values in the set $\{0.1, 0.5, 1.0, 2.0\}$. Preliminary experiments on CIFAR-100-LT (IR=100) suggested the following as strong starting points, which were then validated or slightly adjusted for other datasets:
                \begin{itemize}
                    \item For MDPR-CoOp: $\lambda_{\text{sem}} = 0.1$.
                    \item For MDPR-MaPLe: $\lambda_{\text{sem}} = 1.0$.
                \end{itemize}
            \item \textbf{Stage 2 (Joint Tuning $\lambda_{\text{pa}}, \lambda_{\text{ka}}, T$):} With the selected $\lambda_{\text{sem}}$, $\lambda_{\text{pa}}$ was tuned from the set $\{0.01, 0.05, 0.1, 0.5, 1.0\}$, $\lambda_{\text{ka}}$ from $\{0.001, 0.005, 0.01, 0.05, 0.1\}$, and the KL distillation temperature $T$ from $\{1.0, 2.0, 5.0\}$.
        \end{enumerate}

    \item \textbf{Loss Weight Warm-up:} The weights for $\mathcal{L}_{\text{sem}}$ and $\mathcal{L}_{\text{ka}}$ (i.e., $\lambda_{\text{sem}}$ and $\lambda_{\text{ka}}$) were linearly warmed up from $0$ to their target tuned values over the first $5$ training epochs. This strategy was found to stabilize early training.
\end{itemize}

% \subsubsection{Inference Settings}
% \label{app:inference_settings}
% \begin{itemize}
%     \item \textbf{Logit Fusion Coefficient ($\beta$):} The coefficient $\beta$ in the logit fusion equation ( $\mathbf{z}_{\text{final}} = (1-\beta)\mathbf{z}_{\text{base}} + \beta\mathbf{z}_{\text{sem}}$) was set to $0.5$ by default for all reported results. This implies an equal contribution from the base VLM's predictions and the MDPR's dynamic semantic pathway predictions.
% \end{itemize}

% Optional: If you have a separate appendix for prompt generation, you can refer to it here.
% \subsubsection{LLM for Prompt Generation}
% \label{app:llm_details_ref}
% Details regarding the LLM used for generating the multi-dimensional prompts and the specific query templates are provided in Appendix~\ref{app:prompt_generation_details}.

\section{Language Model Selection for Prompt Generation}
\label{appendix:llm_selection_brief}
\tcbset{colback = red!3!white, colframe = red!75!black}
% \begin{tcolorbox}[colback=red!3!white,breakable, label=Personality Traits]
% \textbf{visual features}: 

% Provide a concise English phrase describing the key visual appearance features of a "\{classname\}".

% Focus on what it looks like (e.g., shape, color, texture, notable parts).

% \end{tcolorbox}

\begin{tcolorbox}[colback=red!3!white,breakable, label=Personality Traits]

\textbf{visual features}: 

Provide a concise English phrase describing the key visual appearance features of a "\{class-name\}".

Focus on what it looks like (e.g., shape, color, texture, notable parts).

The phrase should be approximately \{target-word-count\} words and suitable to complete the sentence: "A \{class-name\} typically appears as \{YOUR PHRASE HERE\}."

Output ONLY the descriptive phrase. Do NOT include "A \{class-name\} typically appears as".

Descriptive phrase for "\{class-name\}":

\textbf{functional-use}: Provide a concise English phrase describing the primary function or purpose of a "\{class-name\}".

Focus on what it is used for.

The phrase should be approximately \{target-word-count\} words and suitable to complete the sentence: "A \{class-name\} is used for [YOUR PHRASE HERE]."

Output ONLY the descriptive phrase. Do NOT include "A \{class-name\} is used for".

Descriptive phrase for "\{class-name\}":

\textbf{contextual-scene}: Provide a concise English phrase describing the common environments or contexts where a "\{class-name\}" is typically found.

Focus on its usual surroundings or scenarios.

The phrase should be approximately \{target-word-count\} words and suitable to complete the sentence: "A \{class-name\} is commonly found in [YOUR PHRASE HERE]."

Output ONLY the descriptive phrase. Do NOT include "A \{class-name\} is commonly found in".

Descriptive phrase for "\{class-name\}":

\textbf{differential-comparison}: Describe the key visual differences of a "\{class-name\}" when compared to a "\{confusing-class-name\}".

Focus on features that distinguish a "\{class-name\}" from a "\{confusing-class-name\}".

The description should be in English, concise, and approximately {target-word-count} words.

Output ONLY the descriptive phrase itself, suitable for completing the sentence: "Unlike a \{confusing-class-name\}, a \{class-name\} has [YOUR PHRASE HERE]."

Output ONLY the descriptive phrase of differences. Do NOT include "Unlike a \{confusing-class-name\}, a \{class-name\} has".

Descriptive phrase of differences for "{class-name}" compared to "{confusing-class-name}":

\textbf{fine-grained-attribute}: Provide a concise English phrase describing one or two highly distinctive or fine-grained visual attributes of a "\{class-name\}" that make it unique or easily identifiable.

Focus on specific, detailed characteristics.

The description should be in English, concise, and approximately {target-word-count} words.

Output ONLY the descriptive phrase itself, suitable for completing the sentence: "A distinctive feature of a \{class-name\} is [YOUR PHRASE HERE]."

Output ONLY the descriptive phrase of the attribute(s). Do NOT include "A distinctive feature of a \{class-name\} is".

Descriptive phrase of attribute(s) for "\{class-name\}":

\end{tcolorbox}

For generating the multi-dimensional semantic prompts required by MDPR, we evaluated several Large Language Models (LLMs), including Qwen2.5, LLaMa4, and DeepSeek-V3. The evaluation primarily considered the statistical properties of the semantic similarities (forming the prior alignment matrix $\mathbf{M}$) between the CLIP-encoded LLM-generated prompts and generic class descriptions. Considering a comprehensive comparison of key metrics (summarized in Table~\ref{tab:llm_similarity_stats_brief_appendix}) and a qualitative assessment of the generated text, we selected \textbf{Qwen2.5} for prompt generation due to its favorable overall performance in semantic alignment and distributional stability.

% Optional: A more condensed table, or integrate into text if space is extremely tight.
\begin{table}[htbp!] % Using [h] to try and keep it close, but it's an appendix.
  \centering
  \small
  \caption{Key statistics for the prior alignment matrix $\mathbf{M}$ (semantic similarities) from prompts by different LLMs on CIFAR-100. }
  \label{tab:llm_similarity_stats_brief_appendix}
  \begin{tabular}{lccc}
    \hline
    \textbf{LLM} & \textbf{Mean} & \textbf{Std} & \textbf{Median} \\
    \hline
    Qwen2.5     & \textbf{0.8371} & \textbf{0.0370} & 0.8452          \\
    LLaMa4      & 0.8360          & 0.0371          & \textbf{0.8457} \\
    DeepSeek-V3 & 0.8354          & 0.0373          & 0.8438          \\
    \hline
  \end{tabular}
  \vspace{-12pt}
\end{table}

\newpage

\appendix
\label{app:ablation}
\begin{figure}[t]
  \includegraphics[width=\linewidth]{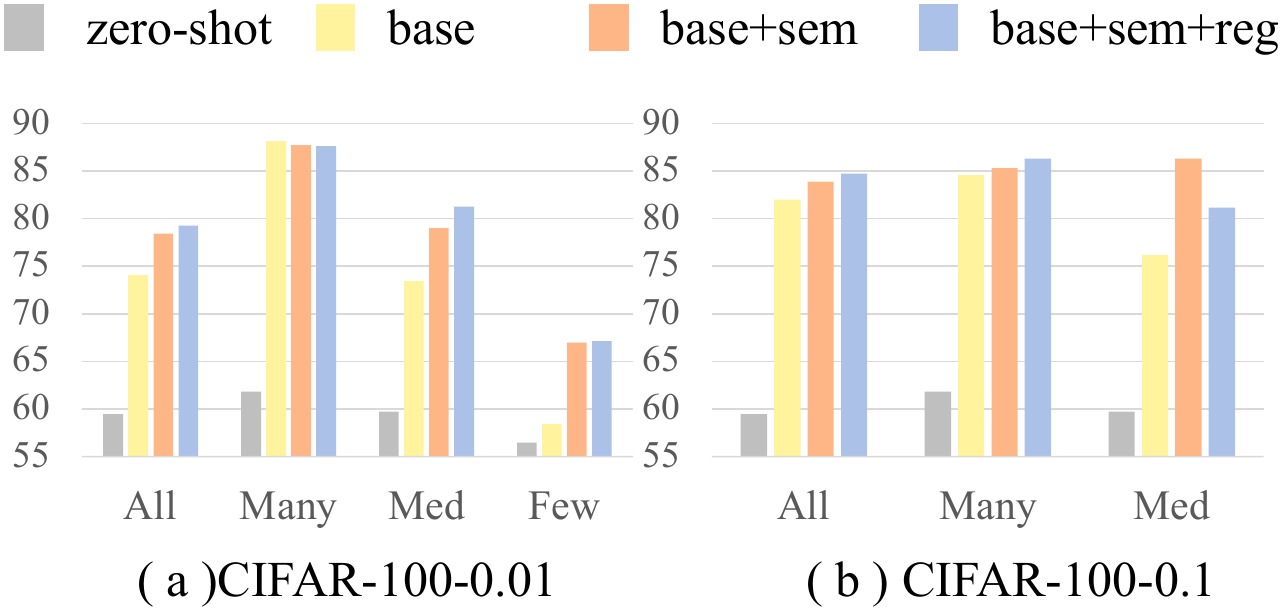}
  \caption{Ablation results across datasets.}
  \label{fig:ablation}
  \vspace{-10pt}
\end{figure}

\begin{algorithm}
\caption{Multi-dimensional Dynamic Prompt Routing (MDPR)}
\begin{algorithmic}[1]
\Require Training set $\mathcal{D}=\{(x_b,y_b)\}_{b=1}^B$
\Ensure Trained model $\phi$
\State Initialize CLIP with pre-trained weights
\State Build confusion matrix $\mathbf{K} \xleftarrow{\text{CLIP}} \mathcal{D}$
\State Generate prompts ${\mathcal{P}_c}\xleftarrow{\text{LLM}}(\mathbf{K},\text{prompts})$
\State Compute $\mathbf{M} \in \mathbb{R}^{C \times 5} \xleftarrow{\text{CLIP}} {\mathcal{P}_c}$
\State Encode $\mathbf{f}_{p} \in \mathbb{R}^{C \times 5 \times d} \xleftarrow{\text{CLIP}} {\mathcal{P}_c}$
\For{\(x_b,y_b\)} in D, \do
    \State Compute $\bf{f}_{ib}=\phi_v(x_b)$
    \State Calculate $\hat{\mathbf{y}}_{cb}$, constrained by $\mathcal{L}_{\text{base}}$
    \State Initialize $\mathbf{f}_{rb}$ and $\mathbf{W}_{r}$
    \For{class $c=1$ to $C$}
        \State $\mathbf{f}^{c}_{rb}, \mathbf{W}^{c}_{r} = \operatorname{C-MHA}
        (\mathbf{f}^{c}_{ib},\mathbf{f}^{c}_{p},\mathbf{f}^{c}_{p})$
        \State Calculate $\mathcal{L}_{\text{reg}}$
        \State Append $\mathbf{f}_{rb}^{c}$ to $\mathbf{f}_{rb}$, $\mathbf{W}^{c}_{r}$ to $\mathbf{W}_{r}$
    \EndFor
    \State Calculate $\hat{\mathbf{y}}_{rb}$, Constraint $\mathcal{L}_{\text{sem}}$
    \State Optimize $\mathcal{L}_{\text{total}}$
    \State Update $\phi \leftarrow \phi - \eta \nabla_{\phi} \mathcal{L}_{\text{total}}$
\EndFor
\State \Return $\phi$
\end{algorithmic}
\label{alg:MDPR}
\end{algorithm}

\end{document}